%% file: PaperForReview.tex
\documentclass[10pt,twocolumn,letterpaper]{article}

\usepackage{iccv}
\usepackage{times}
\usepackage{epsfig}
\usepackage{graphicx}
\usepackage{amsmath}
\usepackage{amssymb}
\usepackage{caption}
\usepackage{subcaption}
\usepackage{multirow}
\usepackage{tikz}

\usepackage[pagebackref=true,breaklinks=true,letterpaper=true,colorlinks,bookmarks=false]{hyperref}
\usepackage[breaklinks=true,bookmarks=false]{hyperref}

\iccvfinalcopy 

\usepackage[capitalize]{cleveref}
\crefname{section}{Sec.}{Secs.}
\Crefname{section}{Section}{Sections}
\Crefname{table}{Table}{Tables}
\crefname{table}{Tab.}{Tabs.}

\def\iccvPaperID{23} 

\begin{document}

\title{Class-Incremental Learning of Plant and Disease Detection: Growing Branches with Knowledge Distillation}

\author{Mathieu Pagé-Fortin\\
Laval University, Canada\\
{\tt\small mathieu.page-fortin.1@ulaval.ca}
}
\maketitle
\begin{abstract}
This paper investigates the problem of class-incremental object detection for agricultural applications where a model needs to learn new plant species and diseases incrementally without forgetting the previously learned ones. We adapt two public datasets to include new categories over time, simulating a more realistic and dynamic scenario. We then compare three class-incremental learning methods that leverage different forms of knowledge distillation to mitigate catastrophic forgetting. Our experiments show that all three methods suffer from catastrophic forgetting, but the Dynamic Y-KD approach, which additionally uses a dynamic architecture that grows new branches to learn new tasks, outperforms ILOD and Faster-ILOD in most settings both on new and old classes.

These results highlight the challenges and opportunities of continual object detection for agricultural applications. In particular, we hypothesize that the large intra-class and small inter-class variability that is typical of plant images exacerbate the difficulty of learning new categories without interfering with previous knowledge. We publicly release our code to encourage future work. \footnote{\url{https://github.com/DynYKD/Continual-Plant-Detection}}  
\end{abstract}

\begin{figure*}
    \centering
\includegraphics[width=0.9\textwidth]{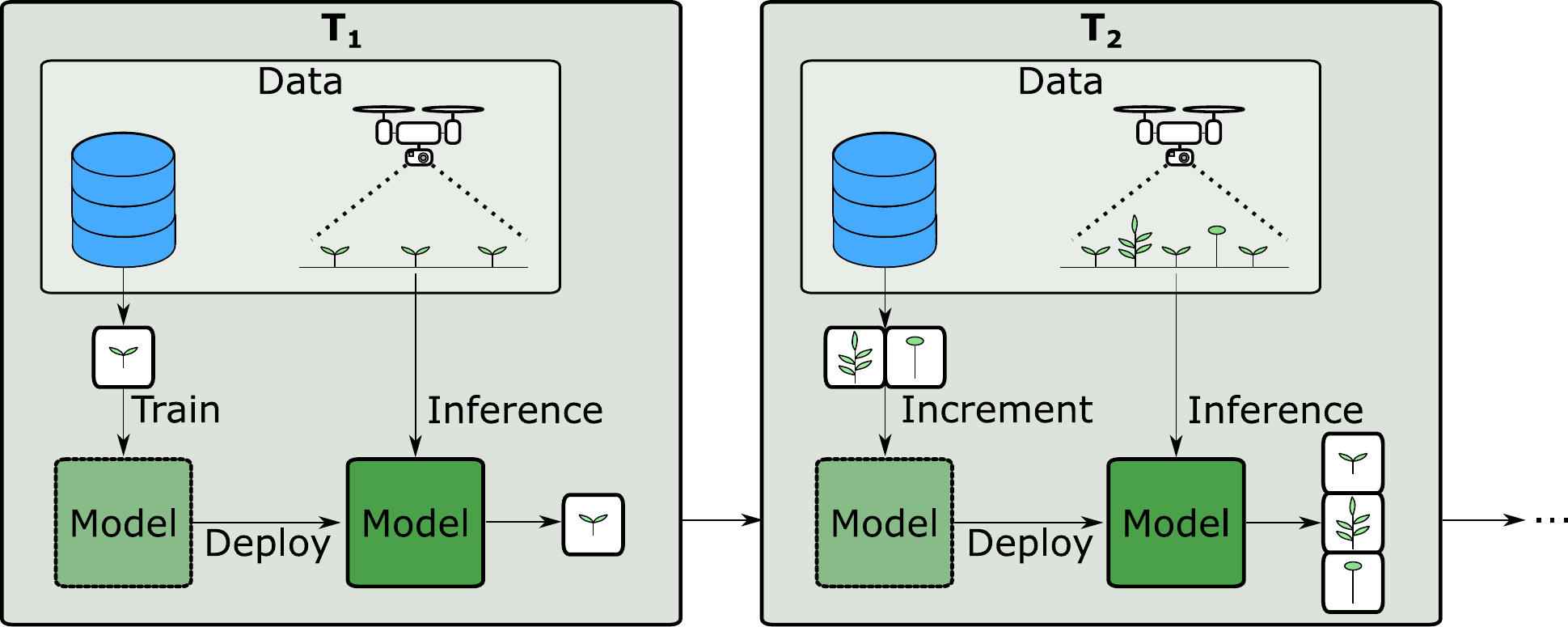}
    \caption{\small Motivating example of continual learning in agricultural contexts. Initially at T$_1$, a detection model is trained to identify a particular species of plant and is subsequently deployed in the field for this purpose. Later at T$_2$, two new weed species emerge in the field. The user intends for the model to accurately detect these additional species as well. To achieve this, the model is incremented exclusively with training examples of these newly emerged weeds and redeployed to detect all three plant varieties. }
    \label{fig:motivating}
\end{figure*}

\section{Introduction}
\label{sec:intro}
Agriculture is a fundamental sector of the global economy, and its importance will only increase as the world's population is expected to reach 9.7 billion by 2050 \cite{unitednations}. With a projected $70\%$ increase in demand for food production, precision agriculture practices are gaining increasing attention to improve productivity by acting at a more refined scale~\cite{shaikh2022towards}. The advent of precision agriculture tools and new technologies has enabled the gathering and analysis of large quantities of data with higher spatio-temporal resolution. Among these advances, computer vision has an important role in remote and proximal sensing by providing tools to process complex images \cite{couliably2022deep, lu2020survey}. 

Remote sensing methods, which include the analysis of data collected from drones, planes and satellites, have many applications for plant health and growth monitoring \cite{patricio2018computer}, yield size and quality estimation \cite{van2020crop}, decision support systems for irrigation or soil management \cite{jha2022irrigation}, large-scale phenotyping \cite{li2020review}, and others. Similarly, proximal sensing methods with personal cameras or cameras mounted on vehicles can be used for insect and pest detection \cite{liu2021plant}, robotic-assisted weed removal or spraying \cite{wu2021review}, and more.

For many of these applications, object detection plays a crucial role in identifying the location of target instances in images, such as individual plants, weeds or insects. To solve this task, deep learning approaches have gained widespread popularity, especially Convolution Neural Networks~(CNN), due to their superior performance and ability to extract relevant features directly from image data \cite{li2021survey}. 

However, deep learning approaches generally assume that the training dataset is static, such that training can be performed in one step before deployment. This scenario faces limitations in dynamic environments \cite{lesort2020continual} typical of agriculture (see Fig. \ref{fig:motivating}). For instance, in the context of crop health monitoring, new diseases or new weed species can emerge on farms by means of contaminated crop seeds \cite{monaco2002weed} or invasive weed species \cite{shabani2020invasive}, inducing a necessity to update existing models to recognize new categories. However, incrementing deep learning models with such new classes is still a challenging task as these models are prone to catastrophic forgetting \cite{mccloskey1989catastrophic}. They tend to erase previous knowledge while learning new classes.

While this problem is actively studied for image classification \cite{de2021continual}, class-incremental object detection is far less explored \cite{menezes2023continual}. Furthermore, previous work has mostly been done on benchmark datasets such as COCO \cite{lin2014microsoft} or Pascal-VOC \cite{everingham2009pascal}. Very few work addressed the challenges of class-incremental learning in agriculture \cite{bhattarai2021incremental, ouadfel2022incremental, li2020ann}. For instance, the large intra-class and small inter-class variability that is characteristic of agricultural imagery (e.g. see Fig.~\ref{fig:SDD} and Fig.~\ref{fig:OPPD}) might increase catastrophic forgetting as new knowledge interferes with previously learned features. Moreover, as the annotation of data often requires specialists, the high cost of data acquisition can result in fewer training examples \cite{blok2022active}. Consequently, incremental methods should exhibit high plasticity to efficiently learn from few data. On the other hand, increased plasticity generally conflicts with stability, such that forgetting can be exacerbated.

In this paper, we study class-incremental object detection on two agricultural datasets. To the best of our knowledge, we are the first to tackle such problem explicitly on agricultural imagery. We adapt two public datasets to simulate class-incremental learning scenarios: 1) the Strawberry Disease Detection Dataset \cite{afzaal2021instance} which is composed of 2,500 images showing seven categories of diseases, and 2) the Open Plant Phenotyping Database \cite{leminen2020open}, a large dataset of plant seedlings from 47 different species. We divide the classes in distinct sets and the models have to learn them successively. We publicly release the code that adapts both datasets to class-incremental learning scenarios to encourage future research.

We compare three continual learning approaches that rely on different forms of knowledge distillation (KD) \cite{hinton2015distilling}, a popular regularization strategy that uses the outputs of the previous model to transfer its knowledge to the learning model. Forgetting is thereby reduced as it constrains the new model from diverging significantly from its previous state. The first approach, ILOD \cite{shmelkov2017incremental}, uses KD by forcing the new model to give scores and boxes of previous classes that are similar to the ones given by the previous model. The second, Faster-ILOD \cite{peng2020faster}, includes additional KD losses to further guide the new model in preserving past activations while learning new categories. Finally, Dynamic Y-KD \cite{dynykd} is a hybrid approach that has been proposed recently. It combines a new form of KD that improves the learning of new categories with a dynamic architecture that grows additional feature extractors for each new task. By using class-specific feature extractors to recognize previous categories, catastrophic forgetting has been shown to be significantly reduced.

We perform extensive experiments on the Strawberry Disease Detection Dataset and the Open Plant Phenotyping Database, and we show that while ILOD and Faster-ILOD performs comparatively, Dynamic Y-KD significantly outperforms both of them on new classes while reducing forgetting in most scenarios. However, Dynamic Y-KD incurs higher memory and computational costs at inference, which can be an issue if real-time inference is critical.  

In summary, our contributions are as follows:

\begin{itemize}
    \item We adapt two datasets of agricultural imagery to study the task of class-incremental object detection. To our knowledge, these are the first public datasets for continual learning of tasks related to agriculture. We publicly release the code to generate the same splits for reproducibility and to encourage future work. 
    \item We compare three continual learning approaches of object detection that exploit knowledge distillation and a dynamic architecture. Notably, we show that the recent Dynamic Y-KD generally outperforms ILOD and Faster-ILOD at a cost of increased computation.
    \item We perform extensive experiments on various class-incremental scenarios on both datasets. Our results reveal that continual learning is especially challenging on agricultural data, most likely because of the high intra-class and low inter-class variability.
\end{itemize}

\section{Related work}

\subsection{Object Detection in Agriculture}
Object detection is a fundamental task for several applications in agriculture, including pest detection \cite{du2022novel, lippi2021yolo, wang2021agripest}, automatic weed spraying \cite{ghatrehsamani2023artificial, salazar2022beyond, abdulsalam2020deep, krishnan2021comprehensive}, harvesting \cite{zhang2020multi}, yield estimation \cite{koirala2019deep, sozzi2021grape, lee2020artificial}, crop health and growth monitoring \cite{song2020automatic, zhang2020deep, cho2023plant}, and more.   

Agricultural contexts exhibit specific challenges that complicate the application of computer vision techniques. For instance, real-time inference can be required for robotic solutions \cite{walia2021methodology, rakhmatulin2021deep, zheng2018real}. The objects of interest (e.g. fruits) can also be small and occluded \cite{wosner2021object}, have irregular and varying shapes \cite{lu2021counting} and can appear under different lighting conditions \cite{zhao2020augmenting}. Furthermore, acquiring and labelling data to train deep learning models is especially costly in agriculture as it can require field experts, for instance to annotate specific diseases and weed species. 

Several deep learning models have been proposed to tackle these challenges. For instance, \cite{wosner2021object} compared three models with regard to their ability to handle images with large numbers of fruits and flowers of varying sizes. The authors of \cite{lu2021counting} proposed an attention mechanism that processes features from multiple resolutions and the use of Atrous Spatial Pyramid Pooling to improve the detection of occluded or small leaves. Regarding the high cost of data annotation, an active learning strategy was proposed in \cite{blok2022active} to only label the most informative images.

However, differently from previous lines of work that aimed to improve accuracy on challenging images, this paper tackles the problem of continual learning on agricultural images.

\subsection{Continual Learning}
While the challenges above have been addressed in previous work, it is generally assumed that the environment is static, such that object detector models can be trained in one step and deployed afterward. Very few work considered the challenges that arise due to the dynamic nature of agricultural environments, such as the emergence of new object categories. For instance, new pests, diseases or weed species can emerge on farms by means of contaminated crop seeds \cite{monaco2002weed} or invasive weed species \cite{shabani2020invasive}. Effective computer vision solutions should therefore be able to be incremented with new data easily. However, deep learning models are prone to catastrophic forgetting, making their adaptation to new classes while preserving previous knowledge especially challenging. 

To mitigate catastrophic forgetting, continual learning solutions generally employ rehearsal \cite{shieh2020continual, verwimp2021rehearsal, maracani2021recall}, regularization losses \cite{kirkpatrick2017overcoming, li2017learning, douillard2021plop, dynykd}, or dynamic architectures~\cite{dynykd, li2018incremental}. Rehearsal consists of replaying past data in training batches while learning new classes, thereby reducing the drift towards new classes. Regularization-based approaches constrain the weights of the learning model to limit knowledge loss. Finally, dynamic architectures keep specific weights for each set of classes and grow new branches to accumulate new knowledge. For a recent and detailed review on continual object detection, please refer to \cite{menezes2023continual}.

Previous work on continual learning in the context of agricultural applications explored the task of plant disease classification \cite{li2020ann, bhattarai2021incremental, ouadfel2022incremental}, mainly using rehearsal strategies. For instance, \cite{li2020ann} uses a Generative adversarial network to generate abstract representations of previous classes to store them in memory and mixes them with new classes. In \cite{bhattarai2021incremental}, raw images of previous tasks are stored and used after incremental learning to correct the bias towards new classes with a linear layer. Finally, \cite{ouadfel2022incremental} employs rehearsal and ensures that diversity of old data is preserved to limit forgetting. 

To our knowledge, regularization strategies and dynamic architectures have not been explored for continual learning to solve tasks related to agriculture. Furthermore, the problem of class-incremental object detection has not been studied on agricultural images in previous work. Therefore, this paper contributes to continual learning in agriculture by comparing three object detection approaches using knowledge distillation and dynamic architectures. 

\begin{figure*}
    \includegraphics[width=\textwidth]{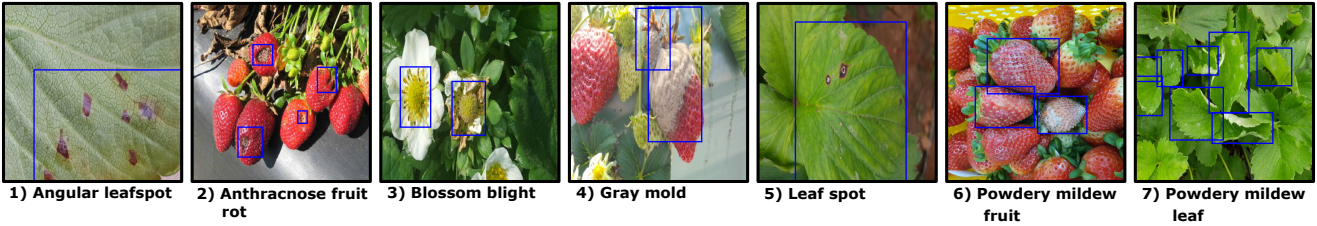}
    \caption{\small Strawberry Disease Detection Dataset \cite{afzaal2021instance}. Examples of images and annotations for the seven categories of diseases.} 
    \label{fig:SDD}
\end{figure*}

\section{Materials and Methods}
In this section, we begin by formulating the problem of continual object detection in the context of agricultural applications. Next, we describe the two agricultural datasets that we used to experiment class-incremental plant and disease detection, and we summarize how we adapted them to simulate incremental learning. We then introduce the three methods that we compared, namely ILOD \cite{shmelkov2017incremental}, Faster-ILOD \cite{peng2020faster} and Dynamic Y-KD \cite{dynykd}.

\subsection{Continual Object Detection in Agriculture}
Continual object detection aims to increment a model~$f_{\theta^{t-1}}$ that can detect objects of classes $\mathcal{C}^{0:t-1}$ to a model $f_{\theta^t}$ that can detect new classes $\mathcal{C}^t$ in addition to previous classes $\mathcal{C}^{0:t-1}$. At any step $t$, we are given a labelled training dataset $\mathcal{D}^t$ composed of images $X^t$ with annotations $Y^t$ showing examples of current classes $\mathcal{C}^t$. For object detection, the annotations $Y^t$ are the bounding boxes and classes of the items on each image. However, the established continual learning setup \cite{shmelkov2017incremental, cermelli2020modeling, dynykd} generally assumes that at any given step, training data are available for current classes exclusively, while previous and future classes are either unlabeled or unavailable. We also adopt this experimental setup in this work.

In the context of agricultural applications, such a continual learning scenario captures a situation where an object detection model $f_{\theta^{t-1}}$ has been trained and deployed to solve a given task, detecting objects (e.g. weeds, diseases) from certain classes $\mathcal{C}^{0:t-1} $ in the field (see Fig. \ref{fig:motivating}). However, as the environment is dynamic, new categories $\mathcal{C}^t$ might emerge at a time $t$, inducing the necessity to update the detection model. After a dataset $\mathcal{D}^t$ of new classes $\mathcal{C}^t$ has been acquired and labeled, the model $f_{\theta^{t-1}}$ is incremented to $f_{\theta^{t}}$. This new model can then be redeployed on the field, having acquired the ability to detect new classes accurately in addition to previous ones. This scenario can repeat for multiple steps if new categories emerge in the future again. 

\begin{figure*}
\centering
    \includegraphics[width=0.6\linewidth]{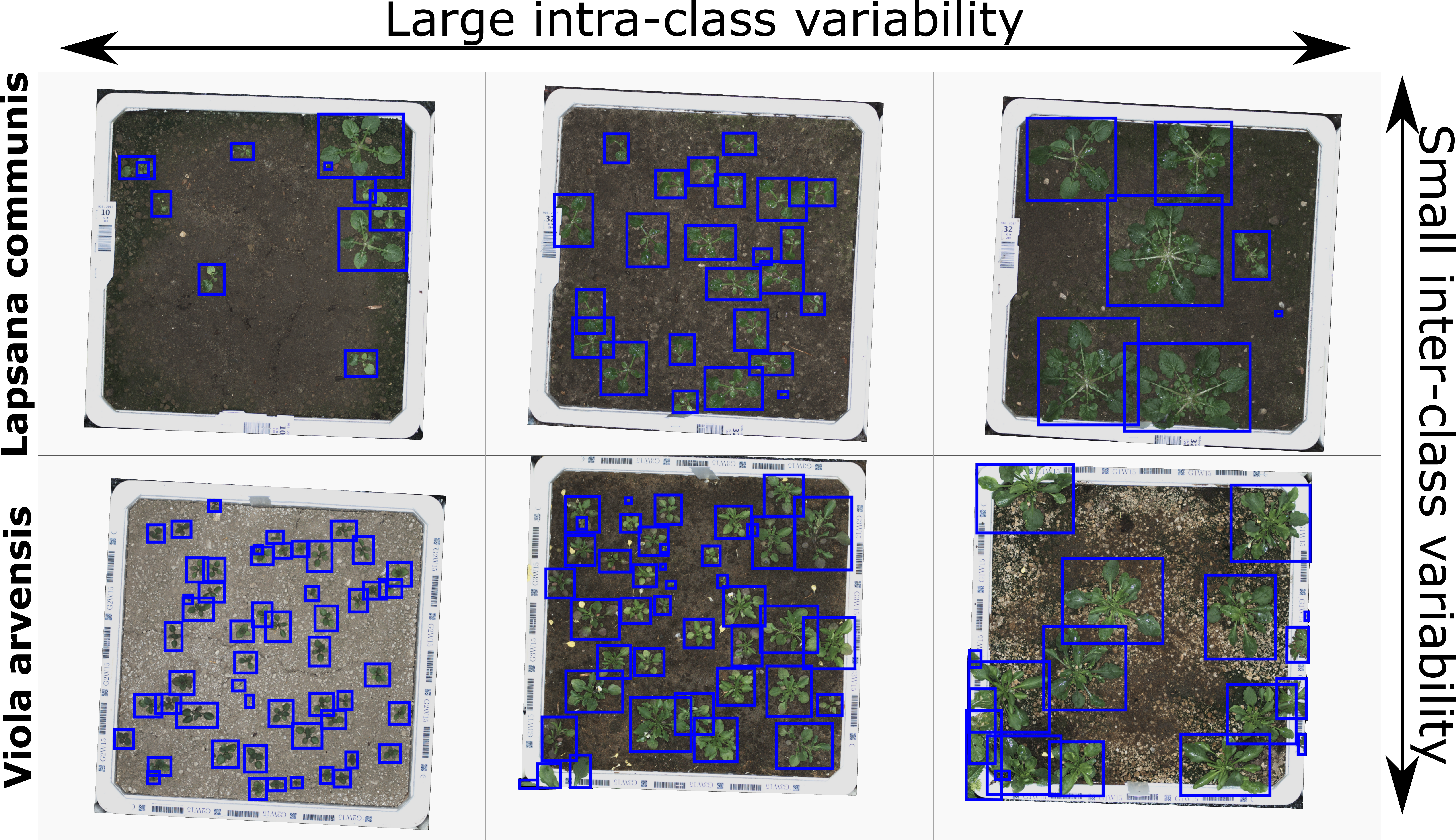}
    \caption{\small Open Plant Phenotyping Database \cite{leminen2020open}. Three examples of two classes and their annotations are shown to highlight the large intra-class variability and small inter-class variability that are frequent in agricultural imagery.}
    \label{fig:OPPD}
\end{figure*}
\subsection{Datasets}
We adapt two public datasets to simulate incremental learning scenarios following the established experimental setups of class-incremental learning \cite{shmelkov2017incremental}. An incremental learning scenario is defined by $b-n$, where $b$ represents the number of base categories learned initially, and $n$ is the number of new classes added in subsequent steps until all classes from the dataset are learned. 

As an example on the Strawberry Disease Detection Dataset \cite{afzaal2021instance}, a scenario denoted by \textit{3-2} designates that the models are trained on categories \textit{1-3} in a first step, then they are incremented exclusively on classes \textit{4-5}, and finally they are trained exclusively on classes \textit{6-7} to reach the total of $7$ categories of diseases. At the end, model performances are evaluated on the test set composed of all classes \textit{1-7}.  

Following the experimental setup established to adapt benchmark datasets such as Pascal-VOC, the classes are simply sorted in alphabetical order. The images from each dataset are split into train and test sets. However, at each incremental step $t$, we only have access to the training set~$\mathcal{D}^t$.

\vspace{0.5em}
\noindent\textbf{Strawberry Disease Detection Dataset (SDDD).} The SDDD \cite{afzaal2021instance} is composed of~2,500 images showing seven types of strawberry diseases. Figure \ref{fig:SDD} illustrates an example of each class. Notice that diseases can be observed either on leaves, flowers or fruits. All images have been acquired with mobile phones and have been resized to a resolution of~$419 \times 419$. This dataset is relatively small for the task, which is a common challenge of public datasets of agricultural imagery \cite{lu2020survey}. For this reason, the authors of SDDD \cite{afzaal2021instance} explored various data augmentation techniques and found that some operations could slightly improve the results over a baseline. For simplicity, in this paper we did not apply such technique but future work should explore the contribution of data augmentation in incremental learning settings.

\vspace{0.5em}
\noindent\textbf{Open Plant Phenotyping Database (OPPD).}
The OPPD~\cite{leminen2020open} regroups 7,590 images of 47 plant species. Each image captures a grow box viewed from the top in which several plants of the same species grow, making a total of 315,041 plant entities. The images have been acquired throughout the whole life cycle of the plants, such that intra-class variability is high while the inter-class variability can be quite small, especially in early stages of development~(see Fig.~\ref{fig:OPPD}).   

\subsection{Methods}
In this work, we compare three approaches to tackle continual object detection. We begin by summarizing the Faster R-CNN model \cite{ren2015faster} since all three approaches build on this architecture. Then, we explain ILOD \cite{shmelkov2017incremental} and Faster-ILOD \cite{peng2020faster}, two methods that exclusively rely on knowledge distillation~(KD)~\cite{hinton2015distilling} to prevent catastrophic forgetting. We summarize the recent Dynamic Y-KD network \cite{dynykd} that uses an hybrid approach between KD and a dynamic architecture.

\subsubsection{Faster R-CNN}
\label{sec:faster-rcnn}
In the context of continual learning, Faster R-CNN \cite{ren2015faster} is made of a backbone $\mathcal{F}_{\theta^t}^B$, a region proposal network~(RPN)~$\mathcal{F}_{\theta^t}^{RPN}$ and a box head $\mathcal{F}_{\theta^t}^{box}$. Specifically, input images $X^t$ are fed to the backbone to give feature maps~$\hat{X}^t$. These features maps are then processed by the RPN which proposes several regions of interests (RoIs) with corresponding objectness scores. The box head subsequently classifies each RoI and the coordinates of bounding boxes are regressed. 

The supervised loss used to train Faster R-CNN is the sum of classification and regression losses on the RPN and the box head. Following the formulation of \cite{cermelli2022modeling}, we define the loss of Faster R-CNN as follows:
\begin{equation}
    \mathcal{L}_{faster} = \mathcal{L}_{cls}^{RPN} + \mathcal{L}_{reg}^{RPN} + \mathcal{L}_{cls}^{box} + \mathcal{L}_{reg}^{box},
\end{equation}
where $\mathcal{L}_{cls}$ are binary cross-entropy losses for classification applied on the RPN and the box head, and $\mathcal{L}_{reg}$ are regression losses of bounding boxes. For more details we refer the reader to \cite{ren2015faster}. 

While this training can be sufficient to learn a single detection task with a closed set of classes, catastrophic forgetting arises when facing incremental learning scenarios. For this reason, class-incremental learning techniques have been proposed to complement Faster R-CNN and address its limitations when new sets of classes are introduced.  

\subsubsection{ILOD}
\label{sec:ILOD}
ILOD \cite{shmelkov2017incremental} is among the first deep learning methods proposed to tackle continual learning for object detection. It employs KD, a popular regularization-based strategy to reduce catastrophic forgetting. To perform incremental learning, ILOD keeps a frozen copy of the whole network from the previous step, acting as a teacher network to a student network. Training images of new classes are then given to both the new and previous models. In addition to the supervised loss, the new model is trained with an L2 loss that compares the difference between classification logits $\hat{y}$ and box coordinates $r$ given by the previous model, as follows:
\begin{equation}
    \mathcal{L}_{dist}^{box} = \frac{1}{N|\mathcal{C}^{0:t}|}\sum \left[(\hat{y}^t - \hat{y}^{t-1})^2 + (r^t - r^{t-1})^2\right],
\label{eq:box_dist}
\end{equation}
where $N$ is the number RoIs sampled for KD. Thereby, this loss encourages the new model to keep similar outputs for previous classes, preventing the weights from diverging from their previous states. 

The total loss of ILOD is then as follows:
\begin{equation}
    \mathcal{L}_{\text{ILOD}} = \mathcal{L}_{faster} + \lambda_1\mathcal{L}_{dist}^{box}, 
\end{equation}
where $\lambda_1$ is a hyperparameter applied on the KD loss to balance the trade-off between learning and forgetting.

\subsubsection{Faster-ILOD}
\label{sec:FILOD}
Faster-ILOD \cite{peng2020faster} adopts a similar approach to continual object detection than ILOD, using KD as the approach to limit forgetting. However, rather than only distilling knowledge from the last layer, namely the class logits and box coordinates, the authors add a KD loss between the feature maps of the teacher and student networks, and another for the RPN. This time, an L1 loss is used for distillation of feature maps activations as follows:~\begin{equation}
    \mathcal{L}_{dist}^{feat} = 
        \frac{1}{HWD}
        \begin{cases}
            \sum_i ||\hat{X}^{t-1}_i - \hat{X}_i^t||_1,&\text{if } \hat{X}^{t-1}_i>\hat{X}_i^t \\
            0, &\text{otherwise}
        \end{cases}
\end{equation}
where $H$, $W$ and $D$ are the height, width and depth of the feature maps, respectively. Notice that KD is only applied if the activation of the previous model is larger than the one of the current model. The motivation is to only distill activations that are important for previous classes.

For the RPN distillation, a similar idea of only distilling the outputs that are most likely to be important for old classes is adopted. Specifically, the distillation between the RPN of the teacher and student networks is as follows:
\begin{align*}
    \mathcal{L}_{dist}^{RPN} = \frac{1}{N}\sum_i^N &\mathbf{1}_{\left [s_i^t\geq s_i^{t-1}\right ]}||s_i^t - s_i^{t-1}|| + \\
    &\mathbf{1}_{\left [ s_i^t \geq s_i^{t-1} + \tau \right]} ||\omega_i^t - \omega_i^{t-1}||, \tag{5}
\label{eq:rpn_dist}
\end{align*}
where $s_i$ is the objectness score of the RoI $i$, $\omega_i$ is its coordinates, and $\tau$ is an hyperparameter that defines a margin.

The total loss of Faster-ILOD is then as follows:
\begin{equation}
    \mathcal{L}_{\text{FILOD}} = \mathcal{L}_{\text{ILOD}} + \lambda_2 \mathcal{L}_{dist}^{feat} + \lambda_3 \mathcal{L}_{dist}^{RPN}.  \tag{6}
\end{equation}

\subsubsection{Dynamic Y-KD Network}
\label{sec:dynykd}
The Dynamic Y-KD network \cite{dynykd} is a recent approach that adopts a hybrid strategy, mixing a novel self-distillation method with a parameter isolation approach that uses previous backbones to predict past classes while growing new branches to accommodate new tasks. 

While most previous work using KD keeps a copy of the complete network, Dynamic Y-KD reduces the teacher network to the detection head. During incremental learning, the backbone is thus shared between the teacher and student network such that the teacher network head receives the same feature maps as the student.

Formally, a backbone $B$ that is fixed at all time is connected to the student feature extractor $F_\theta^t$ to produce feature maps $\hat{X}^t$ as follows:
\begin{align*}
    \hat{X}^t = F^t(B(X^t)). \tag{7}
\end{align*}
The same feature maps are then given to the teacher detection head $H^{t-1}$ whose weights are frozen, as well as to the trainable student head $H^t$ as follows:
\begin{align*}
    \hat{Y}^{t-1} &= H^{t-1}(\hat{X}^t)\\
    \hat{Y}^{t} &= H^{t}(\hat{X}^t). \tag{8}
\end{align*}

KD is then applied with $\hat{Y}^{t-1}$ and $\hat{Y}^{t}$ according to Eq.~\ref{eq:box_dist} and Eq.~\ref{eq:rpn_dist}. The difference resides in the usage of shared feature maps to produce both outputs. With this modification to the general architecture of KD, the teacher network is made more plastic as its feature extractor is also trained with new data. The student model benefits from this plasticity since only the head is constrained to preserve similar outputs. While the architecture used for KD is different, the total loss of Dynamic Y-KD is similar as Faster-ILOD, except for the feature maps distillation since these are shared between the teacher and student networks. The loss for Dynamic Y-KD is as follows:\begin{equation}
    \mathcal{L}_{\text{DynYKD}} = \mathcal{L}_{faster} + \lambda_1 \mathcal{L}_{dist}^{box} + \lambda_3 \mathcal{L}_{dist}^{RPN}. \tag{9}
\end{equation}

 While more plasticity induces more forgetting, the ability to extract discriminative features of previous classes is preserved by reusing the feature extractors of previous steps. At deployment, Dynamic Y-KD uses all backbones that have been trained in previous steps and a merging mechanism combines the predictions made from the feature maps of each backbones. Interestingly, the results in~\cite{dynykd} showed empirically that KD allows the head to still be compatible with previous features, although it has been exclusively trained for thousands of iterations on new classes. The main drawback of this method is that the model grows in parameters and computational costs for each new set of classes learned incrementally.

\input{strawberry_single_step}
\input{strawberry_multi_steps}

\section{Experimental Setup}

\subsection{Evaluation Metrics}

We compare the implemented methods using the mean average precision (mAP) at an intersection over union~(IoU) threshold of $0.5$, denoted mAP@0.5, or averaged over $10$ thresholds spanning from $0.5$ to $0.95$ -- termed as mAP@(0.5, 0.95). All results are measured after the final task of class-incremental learning. Following previous evaluation conventions in class-incremental learning~\cite{cermelli2022modeling, dynykd}, we evaluate:

\begin{itemize}
    \item Stability with mAP averaged over \textit{base} classes (e.g.~classes \textit{1-42} in the \textit{42-5} scenario of OPPD);
    \item Plasticity with mAP averaged over \textit{new} classes (e.g.~classes \textit{43-47} in the \textit{42-5} scenario of OPPD);
    \item Global performance with mAP averaged over \textit{all} classes;
    \item Stability-Plasticity balance with mAP averaged over \textit{intermediate} classes in multi-steps incremental learning scenarios (e.g. classes \textit{2-6} in a \textit{1-1} scenario that has a total of $7$ classes).
\end{itemize}

Finally, we also introduce the mAP ratio with a joint training approach to discuss the degree to which class-incremental methods differ from the non-incremental upper-bound, namely joint training. 

\subsection{Implementation details}
The experiments are conducted with the framework developed in \cite{dynykd}. We use Faster R-CNN with a ResNet-50 backbone initialized from ImageNet for all approaches. In addition to ILOD, Faster-ILOD and Dynamic Y-KD, we compare a naive fine-tuning approach as a baseline that does not use any mechanism to mitigate forgetting, and joint training which learns all classes simultaneously. The latter approach can be seen as an upper-bound that helps to evaluate the ability of continual learning models to tackle class-incremental challenges. In the following we describe the hyperparameters used by all methods for each dataset.

\paragraph{SDDD.} For this dataset, we train each model for 10,000 iterations with SGD with a learning rate of 0.005 on the first set of classes. The learning rate is decayed by a factor of 0.1 at steps 8,000 and 9,500. For the next steps, the models are trained with a learning rate of 0.0001 for 2,500 iterations per class added. The batch size is set to 4 in all setups. We use the same split established in \cite{afzaal2021instance}, which divides the 2,500 images into sets: 1,450 for training, 307 for validation, and 743 for testing. However, due to the small number of images per class, we merge the training and validation splits together to form a larger training set.

\paragraph{OPPD.} We train each model for 50,000 iterations with SGD with a learning rate of 0.005 on the first set of classes. For the next steps, the models are incremented for 10,000 iterations with a learning rate of $0.001$. The batch size is also set to 4 in all setups. We divide the train and test splits by sampling a disjoint set of grow boxes for each species with their corresponding images. Similar to the previous dataset, we adapt the OPPD to incremental scenarios by sorting the plant species in alphabetical order and defining $b-n$ scenarios.

\input{OPPD_single_step}
\section{Results}
\subsection{Strawberry Disease Detection Dataset}
\noindent\textbf{Single-step incremental learning.} We show in Table~\ref{tab:st_strawberry} the results for single-step incremental learning on the SDDD. We can see that catastrophic forgetting is particularly severe in \textit{5-2}, \textit{4-3} and \textit{3-4} as the mAP of all approaches on old classes is less than half of the one obtained by joint training. Similarly, learning new classes incrementally is far from achieving the performance of joint training.

Nonetheless, we can see that Dynamic Y-KD outperforms ILOD and Faster ILOD in most settings both on previous and new classes. For instance, in \textit{6-1}, Dynamic Y-KD obtains similar mAP on classes \textit{1-6}, with $40.1\%$ compared to $39.7\%$ and $39.9\%$ obtained by ILOD and Faster ILOD, respectively. But on the seventh category (\textit{Powdery mildew leaf}), Dynamic Y-KD outperforms ILOD and Faster ILOD by $+2.2\%$ and $+2.9\%$, respectively. In \textit{5-2}, \textit{3-4} and \textit{3-4}, the gap of mAP on new classes is even greater. These results confirm that the self-distillation employed by Dynamic Y-KD facilitates learning of new classes, while the dynamic architecture generally improves performance on old classes. On all classes (\textit{1-7}), Dynamic Y-KD outperforms other methods in every scenario.

\noindent\textbf{Multi-steps incremental learning.} Multi-steps incremental learning represents a more ambitious goal of class-incremental learning as it involves the ability to learn and remember several tasks successively. The results for various multi-steps incremental learning scenarios of the SDDD are shown in in Table \ref{tab:mt_strawberry}.

We can see that forgetting is exacerbated and learning new classes is also complicated in these setups. In the \textit{3-2} and \textit{3-1} scenarios, all three continual learning methods perform similarly on the first set of classes. However, while ILOD and Faster ILOD also perform similarly on new classes, Dynamic Y-KD significantly outperforms them. For instance, in \textit{3-2}, Dynamic Y-KD obtains $+3.4\%$ and $+7.0\%$ mAP on both sets of novel classes compared to the second best approaches. In \textit{3-1}, Dynamic Y-KD outperforms ILOD by $+12.6\%$ mAP on the last class while suffering less from forgetting. 

However, the results are very far from joint training, especially in the \textit{1-1} scenario where the best performances only reach $\frac{9.1\%}{47.5\%}=19.2\%$ of the global performance of joint training. This highlights the difficulty of class-incremental learning, especially on difficult tasks involving agricultural imagery. While current continual learning methods can have reasonable performances in single-step incremental learning, there is a lot of room for improvement on long-term incremental learning.

\subsubsection{Open Plant Phenotyping Database}
We now report the mAP@0.5\% of single-step incremental learning scenarios on the OPPD in Table \ref{tab:st_oppd}. We can observe from the results that this dataset is very challenging. Indeed, the OPPD contains many classes, large amounts of instances on each image, and the plants have large intra-class and small inter-class variability (see Fig. \ref{fig:OPPD}). Similar to the results on SDDD, ILOD and Faster-ILOD perform comparatively in all three setups, while Dynamic Y-KD obtains better results. Interestingly, Dynamic Y-KD reaches similar mAP@0.5\% than fine-tuning, while still suffering less from forgetting than other methods. Indeed, Dynamic Y-KD obtains $22.5\%$, $21.8\%$ and $17.1\%$ compared to $22.8\%$, $22.4\%$ and $19.8\%$ by fine-tuning on new classes in \textit{42-5}, \textit{37-10} and \textit{27-20} scenarios, respectively. 

Overall, Dynamic Y-KD reaches $\frac{21.7}{24.5}=88.6\%$ of the joint training performance on all classes in \textit{42-5}. When more classes are involved, Dynamic Y-KD only reaches $\frac{18.4}{24.5}=75.1\%$ and $\frac{17.3}{24.5}=70.6\%$ of the performance obtained by joint training in \textit{37-10} and \textit{27-20}, respectively.

\subsection{Discussion}
Together, the results of class-incremental learning methods compared to joint training highlight the difficulty of such scenario on agricultural imagery. For instance, on the Strawberry Disease Detection Dataset in the \textit{5-2} scenario, the best-performing method obtains $22.0\%$ on classes \textit{1-5} whereas joint training obtains $46.8\%$ (see Tab. \ref{tab:st_strawberry}), indicating a strong effect of forgetting. 

Future work should investigate the underlying reasons. We hypothesize that the small inter-class variability is especially challenging for class-incremental learning. Notably, as classes $4$ (Gray mold) and $6$ (Powdery mildew) are relatively similar (see Fig. \ref{fig:SDD}) in this dataset, features that are learned to detect the $6^{th}$ class might strongly interfere with those that were used to detect the $4^{th}$ class learned in a previous step. Consequently, the incremental model might learn to suppress previous knowledge to reduce interference and thereby optimize the current task.

Future work should also investigate a combination of rehearsal techniques and other continual learning strategies. For instance, the hybrid approach of Dynamic Y-KD combined with a small memory bank of previous examples might allow preserving more discriminative features of previous classes and reduce forgetting.

\section{Conclusion}
This work is among the first to tackle continual object detection for tasks related to agriculture. To this aim, we adapted two public datasets to simulate incremental learning scenarios in which classes are learned in two or more distinct steps. In each step, the models only have access to current classes for training. 

To address the problem of catastrophic forgetting faced by deep learning models, we compared ILOD, Faster-ILOD and Dynamic Y-KD. All three approaches leverage different forms of knowledge distillation to reduce knowledge loss during incremental learning. Dynamic Y-KD reduces the constraining factor of knowledge distillation by sharing the same backbone between the teacher and student networks, which proved to be more effective at learning new classes on both datasets. Moreover, as Dynamic Y-KD uses a parameter isolation strategy in which previous and new backbones are used at inference, catastrophic forgetting is also mitigated at a cost of increased computation. 

The experiments conducted in this work showed that the three continual learning methods effectively addressed the issue of forgetting while learning new classes, yet their performances are far from the joint training upper-bound. This highlights the potential for further improvements in the field of continual object detection, particularly in the context of agricultural imagery. 

We hypothesized that the inherent challenges posed by the small intra-class and large inter-class variability characteristic of plant images exacerbate the problem of catastrophic forgetting. Future work should explore rehearsal approaches to reduce forgetting, combined with hybrid strategies such as Dynamic Y-KD to improve learning of new classes.


\section*{Acknowledgment}
This work was granted access to the HPC resources of the Digital Research Alliance of Canada.

{\small
\bibliographystyle{ieee_fullname}
\bibliography{egbib}
}

\end{document}

%% file: strawberry_single_step.tex
\begin{table*}[t]
    \centering
    \begin{tabular}{c|ccc|ccc|ccc|ccc}
    \hline
    \multicolumn{1}{c|}{}&\multicolumn{3}{c|}{\textbf{6-1}} & \multicolumn{3}{c|}{\textbf{5-2}}&\multicolumn{3}{c|}{\textbf{4-3}}& \multicolumn{3}{c}{\textbf{3-4}}\\
    \textbf{Method} & \textbf{1-6} & \textbf{7} &\textbf{1-7} &\textbf{1-5} & \textbf{6-7} &\textbf{1-7} & \textbf{1-4} & \textbf{5-7} &\textbf{1-7} & \textbf{1-3} & \textbf{4-7} &\textbf{1-7}\\
    \hline
    Fine-tuning& 17.4 & 35.8 & 20.0 & 10.2 & 35.2 & 17.3&6.9&33.3&18.2&12.7&20.3&17.0\\
    \hline
    ILOD \cite{shmelkov2017incremental} &39.7 & 29.6 & 38.3 & 20.0 & 27.5 & 22.2 &22.4&22.2&22.3&22.9&15.9&18.9\\
    Faster ILOD \cite{peng2020faster} & 39.9 & 28.9 & 38.4 & 20.7& 26.6 & 22.4&21.7&22.3&21.9&\textbf{23.9}&16.0&19.4\\
    Dynamic Y-KD \cite{dynykd}& \textbf{40.1} & \textbf{31.8} & \textbf{38.9} &\textbf{22.0}&\textbf{32.8} & \textbf{25.1}&\textbf{25.2}&\textbf{26.9}&\textbf{25.9} & 21.5&\textbf{20.4}&\textbf{20.9}\\
    \hline
    Joint Training& 45.4 & 60.4 & 47.5 & 46.8 & 49.3 & 47.5 &47.0&48.2&47.5  & 47.5&47.5&47.5\\
    \hline
    \end{tabular}
    \caption{\small mAP@(0.5, 0.95)\% results of single-step incremental instance segmentation on the Strawberry Disease Detection Dataset.}
    \label{tab:st_strawberry}
\end{table*}

%% file: strawberry_multi_steps.tex
\begin{table*}[t]
    \centering
    \begin{tabular}{c|cccc|cccc|cccc}
    \hline
    \multicolumn{1}{c|}{}&\multicolumn{4}{c|}{\textbf{3-2}} & \multicolumn{4}{c|}{\textbf{3-1}}& \multicolumn{4}{c}{\textbf{1-1}}\\
    \textbf{Method} & \textbf{1-3} & \textbf{4-5} & \textbf{6-7}&\textbf{1-7} &\textbf{1-3} & \textbf{4-6}&\textbf{7} &\textbf{1-7} &\textbf{1} & \textbf{2-6} & \textbf{7}&\textbf{1-7}\\
    \hline
    Fine-tuning &0.0&0.0&31.2&8.9&0.0& 0.0 &15.5&2.2&0.0 & 0.0 & 19.4 & 2.8\\
    \hline
    ILOD \cite{shmelkov2017incremental} &\textbf{23.5}&13.9&13.7&17.9& 26.8&8.9 &7.0&16.3&\textbf{13.9}&\textbf{8.0}&10.0&9.1\\
    Faster ILOD \cite{peng2020faster} &22.8&13.4&14.3&17.7&25.3 &8.9 &6.9 &15.7&16.5& 2.7 &6.4 &5.2\\
    Dynamic Y-KD \cite{dynykd} & 23.3 &\textbf{17.3} &\textbf{21.3}&\textbf{21.0}& \textbf{27.0} &\textbf{12.2}&\textbf{19.6}&\textbf{19.6}& 13.0 & 7.1 &\textbf{14.9} &\textbf{9.1}\\
    \hline
    Joint Training& 47.5&45.7&49.3&47.5&47.5&43.3 &60.4 &47.5 &  44.7&45.5 & 60.4 & 47.5\\
    \hline
    \end{tabular}
    \caption{\small mAP@(0.5, 0.95)\% results of multi-step incremental instance segmentation on the Strawberry Disease Detection Dataset.}
    \label{tab:mt_strawberry}
\end{table*}

%% file: OPPD_single_step.tex
\begin{table*}[t]
    \centering
    \begin{tabular}{c|ccc|ccc|ccc}
    \hline
    \multicolumn{1}{c|}{}&\multicolumn{3}{c|}{\textbf{42-5}} & \multicolumn{3}{c|}{\textbf{37-10}}&\multicolumn{3}{c}{\textbf{27-20}}\\
    \textbf{Method} & \textbf{1-42} & \textbf{43-47} &\textbf{1-47} &\textbf{1-37} & \textbf{38-47} &\textbf{1-47} & \textbf{1-27} & \textbf{28-47} &\textbf{1-47}\\
    \hline
    Fine-tuning& 5.6 & 22.8 & 7.5 & 2.5 & 22.4 & 6.8&3.6&19.8&10.6\\
    \hline
    ILOD \cite{shmelkov2017incremental} &19.5 & 20.9 & 19.6 & 17.3 & 20.0 & 17.9 &16.7&15.1&16.0\\
    Faster ILOD \cite{peng2020faster} & 19.5 & 20.8 & 19.6 & \textbf{17.5}&19.8 & 18.0&16.7&15.1&16.0\\
    Dynamic Y-KD \cite{dynykd}& \textbf{21.6} & \textbf{22.5} & \textbf{21.7} &\textbf{17.5}&\textbf{21.8} & \textbf{18.4}&\textbf{17.5}&\textbf{17.1}&\textbf{17.3}\\
    \hline
    Joint Training& 23.9 & 29.0 & 24.5 &23.6& 27.7 & 24.5 &24.3&24.8&24.5
    \\
    \hline
    \end{tabular}
    \caption{\small mAP@0.5\% results of single-step incremental instance segmentation on the Open Plant Phenotyping Database.}
    \label{tab:st_oppd}
\end{table*}